\documentclass[10pt,a4paper]{article}
\usepackage[utf8]{inputenc}
\usepackage[T1]{fontenc}
\usepackage{bm}
\usepackage{authblk}
\usepackage[tbtags]{amsmath}
\usepackage{amssymb}
\usepackage{amsthm}
\usepackage{graphicx}
\usepackage{algorithm}
\usepackage{algpseudocode}

\algblockdefx{MRepeat}{EndRepeat}{\textbf{repeat}}{}
\algnotext{EndRepeat}

\usepackage{appendix}
\usepackage{lastpage}
\usepackage{svg}
\usepackage{blindtext}
\usepackage{subcaption}
\usepackage[preprint]{jmlr2e}
\usepackage{comment}
\usepackage[textsize=tiny]{todonotes}

\newcommand{\NCLASSES}{K}
\newcommand{\NFEATS}[2]{F_{#1#2}}
\newcommand{\NFEATURES}{F}
\newcommand{\NTIMES}[2]{T_{#1#2}}
\newcommand{\NALLTIMES}{T}
\newcommand{\NSAMPLES}[1]{m_{#1}}
\newcommand{\Y}[2]{Y_{#1#2}}
\newcommand{\MEAN}[1]{\mu_{#1}}
\newcommand{\COV}{\Sigma}
\newcommand{\NDIMS}{d}
\newcommand{\rR}{\mathbb{R}}
\newcommand{\nN}{\mathbb{N}}
\newcommand{\deff}{\triangleq}
\newcommand{\Prob}{\mathbb{P}}
\newcommand{\C}[1]{\mathcal{C}_{#1}}
\newcommand{\ie}{i.e.~ }
\newcommand{\TIME}[3]{t^{#1 #2}_{#3}}
\newcommand{\FEAT}[3]{f^{#1 #2}_{#3}}

\DeclareMathOperator*{\argmax}{arg\,max}

\ShortHeadings{Multivariate Functional LDA}{Bordoloi, Réda, Trautmann, Bej and Wolkenhauer}

\jmlrheading{}{2024}{}{02/24}{}{}{Rahul Bordoloi, Clémence Réda, Orell Trautmann, Saptarshi Bej and Olaf Wolkenhauer}

\title{Multivariate Functional Linear Discriminant Analysis for the %
Classification of Short Time Series with Missing Data}
\author[1]{Rahul Bordoloi}
\author[1]{Clémence Réda}
\author[1]{Orell Trautmann}
\author[2]{Saptarshi Bej}
\author[1,3,4,$\dagger$]{Olaf Wolkenhauer}
\editor{Action Editor's name here}
\affil[1]{Institute of Computer Science, University of Rostock, Germany}
\affil[2]{Indian Institute of Science Education and Research, Thiruvananthapuram}
\affil[3]{Leibniz--Institute for Food Systems Biology, Technical University of Munich, Freising, Germany}
\affil[4]{Stellenbosch Institute of Advanced Studies (STIAS), South Africa}
\affil[$\dagger$]{\textbf{Corresponding author:} \texttt{olaf.wolkenhauer@uni-rostock.de}}

\begin{document}
	
    \maketitle
	
    \begin{abstract}%
        Functional linear discriminant analysis (FLDA) is a powerful tool that extends LDA--mediated multiclass classification and dimension reduction to univariate time--series functions. However, in the age of large multivariate and incomplete data, statistical dependencies between features must %
        be estimated in a computationally tractable way, while also dealing with missing data.
        We here develop a multivariate version of FLDA (MUDRA) to tackle this issue and describe an efficient expectation/conditional--maximization (ECM) algorithm to infer its parameters. We assess its predictive power on the ``Articulary Word Recognition'' data set and show its improvement over the state--of--the--art, especially in the case of missing data. MUDRA allows interpretable classification of data sets with large proportions of missing data, which will be particularly useful for medical or psychological data sets.
	\end{abstract}


\section{Introduction}

Linear discriminant analysis (LDA) is a popular strategy still widely used~\citep{zhu2022neighborhood,mclaughlin2023fedlda,graf2024comparing} for classification and dimensionality reduction. The origins trace back to Fisher's discriminant~\citep{fisher1936use} and its multiclass extension~\citep{rao1948utilization}. The main assumption of LDA is that samples of class $i=1,2,\dots,\NCLASSES$ follow a multivariate normal distribution of mean $\MEAN{i} \in \rR^\NDIMS$ and shared covariance matrix $\COV \in \rR^{\NDIMS \times \NDIMS}$ of full rank, denoted $\mathcal{N}_{\NDIMS}(\MEAN{i}, \COV)$. Then the Bayes optimal rule for classification is applied to predict the class $i(x)$ of a new sample $x \in \rR^\NDIMS$ 
\begin{eqnarray}\label{eq:LDA}
i(x) & & \deff \argmax_{i \in \{1,2,\dots,\NCLASSES\}} \Prob(C=i | X=x)\;,\\
\text{and } & & \Prob(C=i | X=x) \propto \mathcal{N}_{\NDIMS}(x; \MEAN{i}, \COV)\pi_i \propto (\MEAN{i}^\top \COV^{-1})x-\frac{1}{2}\MEAN{i}^\top \COV \MEAN{i}+\log \pi_i \;, \nonumber
\end{eqnarray}
where $\pi_i  \deff \Prob(C=i)$ is the prior probability of class $i$, by applying the Bayes theorem, reordering and ignoring terms that are constant across classes. This classification rule leads to a decision function linear in $x$. Parameters $(\MEAN{i})_{i \leq \NCLASSES}$, $\COV$, $(\pi_i)_{i \leq \NCLASSES}$ can be inferred by maximizing the joint likelihood on a set of training samples.

The advance of functional linear discriminant analysis (FLDA)~\citep{james_functional_2001} allows us to handle cases where $x$ is not simply a $\NDIMS$--dimensional vector, but the output of a time--dependent multivariate function $f$ associated with sample $s$: $x_s=f_s(t) \in \rR^{\NDIMS}$. Note that FLDA is restricted to the univariate case, \ie when $\NDIMS=1$. In practice, one has of course no access to the true function $f_s : \nN^* \rightarrow \rR^{\NDIMS}$, but only to some of its values $f_s(t^s_1), f_s(t^s_2), \dots, f_s(t^s_{n_s})$ at a small number $n_s$ of time points $t^s_1$, $t^s_2$, $\dots$, $t^s_{n_s}$. Such short time series appear in many practical scenarios. %
\begin{figure}[H]%
    \centering
    \includesvg[width=0.76\textwidth]{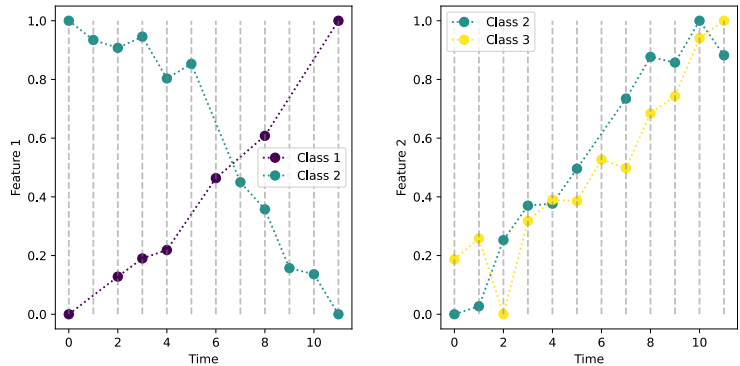}
    \caption{Example of a short multivariate time series with irregular sampling intervals and %
    missing data. Each color %
    corresponds to a class%
    , and each plot represents a single %
    feature. Some features or time points might be missing across individuals, for instance, Feature 2 is not measured for the individual in Class 1, and Feature 1 is not measured at the same time points between Class 1 and Class 2.} %
    \label{fig:demoMissing}
\end{figure}
For instance, think of medical check--ups or psychological studies, where patients (individuals) from different conditions (classes) %
are surveyed for several variables (features) sparsely over weeks or months. 
Ignoring that those values come from the same function and simply concatenating values across all time points will lead to high--dimensional data strongly correlated across time points. This, in turn, would lead to numerical issues that prevent the estimation of the covariance matrix $\Sigma$.

Another challenge is that sampling occurs at irregular, short intervals of time points across samples (\ie different, small values of $n_s$ and different $(t^s_i)_i$ across samples $s$). Moreover, different sets of features might be measured in two distinct samples, meaning that even the value of $\NDIMS$ can vary across samples. Those challenges are exemplified in Figure~\ref{fig:demoMissing}. %
Notably, in this example, there is substantial missing data, such as Feature $2$ in Class $1$ and Feature $1$ in Class $2$, making traditional padding and imputation approaches impractical \citep{bier_variable-length_2022}.
\subsection{Related Work} 

In functional data analysis~\citep{yao_functional_2005}, time--series observations from sample $s$ are considered as noisy samples from the underlying function $f_s$. This approach is particularly promising when one is interested in the overall shape of data. For example, for medical longitudinal data \citep{wang_systematic_2022} or data arising from psychological studies \citep{jebb_time_2015}, one is more interested in trends or the overall pattern of the entire cohort, rather than in individual fluctuations and noise sources. In many practical scenarios, the training data is collected from a large number of subjects at a small number of time points \citep{yoon_estimating_2019}. %
The grouping of subjects is based on a similarity of the curves or trends displayed. The learned model can then be used as a classifier to predict class membership for data from new subjects not in the training data set. In such cases, conventional LDA as shown in Equation~\ref{eq:LDA} can't be directly applied, and time interval discretization is required.

\cite{james_functional_2001} addressed this issue by proposing a model which approximates observations as a noisy linear combination of spline curves, and applies a LDA--like inference procedure to the linear combination coefficients. However, this model only holds for univariate functional data. %
Applying FLDA individually to each curve is suboptimal due to the high dimension of data points %
and the fact that FLDA is not designed to estimate inter--feature correlations.

Multivariate time--series classification (TSC) methods have also been developed in the last decade, for instance, the autoregressive integrated moving average applied to multivariate data (marima)~\citep{spliid_multivariate_2016}. However, marima is only suitable for long--time series. The highest performing approaches for short TSC are today ensemble--of--classifiers methods, such as HIVE--COTE 1.0~\citep{lines_hive-cote_2016} and 2.0~\citep{middlehurst2021hive}, ROCKET~\citep{dempster_rocket_2020} and one of its most recent extension MultiRocket~\citep{tan2022multirocket}. HIVE--COTE algorithms combine the class probabilities returned by four different types of classifiers through a weighting step, which outputs the consensus probabilities, whereas ROCKET--like approaches apply a large number of data transformations using convolutional neural networks to create a massive amount of features. A linear classifier is then trained on those transformed features. %
Nonetheless, the ensemble architecture of HIVE--COTE and ROCKET might hinder the interpretability of the class probability outputs. Moreover, ROCKET--like methods require padding or imputation for handling missing data, which worsens the issue of high dimensional points and impedes interpretability again.

Statistical models tailored to multivariate functional data might overcome the issue of interpretability. Similarly to our work, \cite{gardner-lubbe_linear_2021} presented an extension of FLDA to multivariate data. However, this approach completely ignores the issue of missing features. %
An alternative method proposed by~\cite{fukuda_multivariate_2023} involves the regularization of the underlying time--dependent functions, under the assumptions of regular sampling times and non--missing features. This thus proves impractical in our use case. %

Finally, another line of research focuses on %
the growth mixture model~\citep{ram_growth_2009}, which describes unobserved subpopulations as a noisy time--dependent linear combination of two latent variables. This approach is indeed %
adept at recognizing trends in longitudinal data, but the original model is also ill--equipped at handling missing data. The literature on the topic only partially solves this problem, as the issue of irregular sampling time points has been tackled in the prior works reviewed %
by~\cite{lee_handling_2023}. Even so, %
the proposed advancements cannot handle multiple, highly correlated and missing features.
\subsection{Contributions}

We introduce a multivariate functional linear discriminant analysis algorithm called MUDRA to tackle these issues. Our approach readily extends the scope of functional linear discriminant analysis described by~\cite{james_functional_2001} to multivariate and missing data. MUDRA only requires that the set of features measured for a fixed sample $s$ is the same across all the time points $t^s_1$, $t^s_2$, $\dots$, $t^s_{n_s}$. %
By approaching the problem from the purview of reduced rank regression, we were able to incorporate both time and feature--based irregularity without %
imputation. MUDRA generates a low--dimensional representation, based on the PARAFAC~\citep{bro_parafac_1997} tensor decomposition algorithm, that is particularly efficient for data sets with missing data and generates interpretable patterns. It can also reconstruct the mean time--dependent function across all samples from a class. We experimentally show the superior predictive power of MUDRA compared to baselines for time--series classification on a synthetic data set and on the ``Articulary Word Recognition'' data set, in the presence of a large amount of missing data and when a low dimension is enough to capture the information from data. When there is a large amount of missing data, a common occurrence in real--life applications, MUDRA consistently outperforms ROCKET for low and high dimensionality. %

\subsection{Notation} 

If $X$ is an $m\times n$ matrix, we will denote the rows and columns of $X$ as $\mathbf{x}_{1,:}, \mathbf{x}_{2,:},\ldots,\mathbf{x}_{m,:}$ and $\mathbf{x}_{:,1}, \mathbf{x}_{:,2},\ldots,\mathbf{x}_{:,n}$ respectively. The $ij^{th}$ element of $X$ will be denoted by $x_{ij}$. If $\bm{x}_t$'s are $m$--dimensional vectors, the $i^{th}$ component of $\bm{x}_t$ is denoted by $x_{t,i}$.  The \textit{column space} of $X$, which is the subspace spanned by $\{\mathbf{x}_1, \mathbf{x}_2,\ldots,\mathbf{x}_n\}$ is denoted by $\mathcal{C}(X)$. A general pseudo--inverse of $X$ is denoted by $X^-$. Recall that if $X$ is square and non--singular, the general pseudo--inverse is equal to the inverse: $X^-=X^{-1}$. We denote the identity matrix I regardless of its dimension.

For an $m\times n$ matrix $X$, $vec(X)$ denotes the \textit{vectorization} of $X$ as an $mn$ dimensional vector, formed by stacking all of the columns of $X$ sequentially, \ie
\begin{equation*}
    vec(X) \deff \begin{bmatrix}
	\mathbf{x}_{:,1}^\intercal  &\mathbf{x}_{:,2}^\intercal \cdots &\mathbf{x}_{:,n}^\intercal 
    \end{bmatrix}^\intercal \;.
\end{equation*}
The \textit{Kronecker product} of two matrices $A_{m\times n}$ and $B_{p\times q}$ is an ${mp\times nq}$ matrix, defined by \[A\otimes B \deff \begin{bmatrix}
	a_{11}B &a_{12}B &\cdots &a_{1n}B\\
	a_{21}B &a_{22}B &\cdots &a_{2n}B\\
	\multicolumn{4}{c}{\vdots}\\
	a_{m1}B &a_{m2}B &\cdots &a_{mn}B\\
    \end{bmatrix}\;.\]We will use Kronecker products extensively to turn %
    tensor computations into simpler matrix computations.
A random $m\times n$ matrix $X$ is said to follow a \textit{matrix--variate normal distribution} \citep{nagar_matrix_1999} with \textit{mean} matrix $M_{m\times n}$, \textit{row covariance} matrix $\Sigma_{m\times m}$ and \textit{column covariance} matrix $\Psi_{n\times n}$, if $vec(X)\sim\mathcal{N}_{mn}(vec(M), \Psi\otimes\Sigma)$. It is denoted by $X\sim\mathcal{N}_{m\times n}(M, \Sigma, \Psi)$.

We will denote the variance-covariance matrix of a vector $\bm{y}$ as $\mathbb{V}(\bm{y})$. For a matrix $\Sigma$, $||\bm{y}||_\Sigma$ will denote the \textit{Mahalanobis norm} of $\bm{y}$ with covariance $\Sigma$, defined as $||\bm{y}||^2_\Sigma \deff \bm{y}^\intercal \Sigma^-\bm{y}$. Similarly, for another vector $\bm{x}$, $\langle\bm{x},\bm{y}\rangle_\Sigma$ will denote the \textit{Mahalanobis inner product} with covariance $\Sigma$, defined as $\langle\bm{x},\bm{y}\rangle_\Sigma \deff \bm{x}^\intercal \Sigma^-\bm{y}$. The Mahalanobis inner product thus considers the correlation and variance between two vectors, which will be useful in finding linearly independent discriminants. %

\section{The MUDRA Algorithm}

Formally, we introduce the available short multivariate time--series training data as follows. We consider a class $\C{i}$ of samples in which, for any sample $j$, $\NFEATS{i}{j}$ features at $\NTIMES{i}{j}$ time points have been measured: features $\FEAT{i}{j}{1}$, $\FEAT{i}{j}{2}$, $\dots$, $\FEAT{i}{j}{\NFEATS{i}{j}} \in \nN^*$ at times $\TIME{i}{j}{1}$, $\TIME{i}{j}{2}$, $\dots$, $\TIME{i}{j}{\NTIMES{i}{j}} \in \nN^*$, for $i=1,2,\dots,\NCLASSES$ and $j \in \C{i}$. The matrix of measurements of dimension $\NTIMES{i}{j} \times \NFEATS{i}{j}$ for sample $j$ in class $i$ is denoted $\Y{i}{j}$. The set of distinct features measured across all samples is denoted $\NFEATURES$, and the set of all sampling time points is $T$, that is
\[ \NFEATURES \deff \bigcup_{i \leq \NCLASSES} \bigcup_{j \in \C{i}} \bigcup_{k \leq \NFEATS{i}{j}} \{ \FEAT{i}{j}{k} \} \quad \text{ and } \quad \NALLTIMES \deff \bigcup_{i \leq \NCLASSES} \bigcup_{j \in \C{i}} \bigcup_{k \leq \NTIMES{i}{j}} \{ \TIME{i}{j}{k} \} \;.\]
Note that, since data %
is sampled irregularly, not all observations have been observed at some common time points or with common features. Thus, all matrices $\Y{i}{j}$ will possibly have different dimensions. %
Finally, the number of samples in class $i$ is $|\C{i}| = \NSAMPLES{i}$. 

In the remainder of this section, we describe a model that extends FLDA~\citep{james_functional_2001} to multivariate functional data, that is, that models the observations as a noisy linear combination of spline functions which incorporates multiple features for the first time. Then, we propose an efficient parameter inference algorithm for this model and illustrate how to use it for classification and dimension reduction.

\subsection{A Novel Multivariate Model for Time--Dependent Functional Observations}

Similarly to the FLDA model introduced by~\cite{james_functional_2001}, we consider $b>0$ B--spline functions of order 3%
~\citep{paul_h_c_eilers_flexible_1996} denoted $s_1, s_2, \dots , s_b$ to generate the space of functionals available to our model in a flexible fashion. 
Then for the %
multivariate observations associated with sample $j$ in class $\C{i}$, its %
spline matrix $S_{ij}$ of dimensions $T_{ij}\times b$ is constructed by evaluating the B--spline functions at each time point $\TIME{i}{j}{k}$ %
\begin{equation}\label{eq:Bspline}
    S_{ij} \deff \begin{bmatrix}
    s_1(\TIME{i}{j}{1}) &s_2(\TIME{i}{j}{1}) &\cdots &s_b(\TIME{i}{j}{1})\\
    s_1(\TIME{i}{j}{2}) &s_2(\TIME{i}{j}{2}) &\cdots &s_b(\TIME{i}{j}{2})\\
    \multicolumn{4}{c}{\vdots}\\
    s_1(\TIME{i}{j}{\NTIMES{i}{j}}) &s_2(\TIME{i}{j}{\NTIMES{i}{j}}) &\cdots &s_b(\TIME{i}{j}{\NTIMES{i}{j}})\\
\end{bmatrix} = \left(s_{l}(\TIME{i}{j}{k})\right)_{k \leq \NTIMES{i}{j}, l \leq b}\:.
\end{equation}
We describe now how our model deviates from FLDA and integrates multiple features. Let $C$ be the $\NFEATURES \times \NFEATURES$ identity matrix. Then, we truncate and reorder %
columns of $C$ depending on the actual %
observed features for each sample. %
For sample $j \in \C{i}$, define $C_{ij}$ as the $F\times F_{ij}$ matrix formed by concatenating %
the $(\FEAT{i}{j}{1})^\text{th}$, $(\FEAT{i}{j}{2})^\text{th}$, $\dots$, $(\FEAT{i}{j}{\NFEATS{i}{j}})^\text{th}$ columns of $C$. %
We also introduce the spline coefficient matrix $\eta_{ij}$ of dimensions %
$b\times \NFEATURES$. %
Similarly to the standard LDA model shown in Equation~\ref{eq:LDA},  
\[\eta_{ij} = \MEAN{i} + \gamma_{ij}\;,\]
where $\MEAN{i} \in \rR^{b \times \NFEATURES}$, $\gamma_{ij}\sim \mathcal{N}_{b\times \NFEATURES}(0,\Sigma,\Psi)$ is the autoregressive noise, and $\Sigma$ and $\Psi$ are both positive definite matrices.
Finally, let $\epsilon_{ij}$ be the random matrix of measurement errors on sample $j \in \C{i}$. We assume that it follows a matrix--variate normal distribution, i.e. $\epsilon_{ij}\sim\mathcal{N}_{T_{ij}\times F_{ij}}(0,\sigma^2I,I)$ for some $\sigma>0$. %
We then propose the following model of the multivariate time--dependent observations from sample $j$ in class $i$
\[Y_{ij} = S_{ij}(\MEAN{i} + \gamma_{ij})C_{ij} + \epsilon_{ij}\;.\]
We could leave our model at that, and go on inferring the values of parameters $\MEAN{i}$ for $i=1,2,\dots,\NCLASSES$, $\sigma$, $\Sigma$, and $\Psi$. However, incorporating a r%
educed--rank regression into the model, by adding a supplementary rank constraint on the model parameters, leads to more interpretable models and easier dimension reduction~\citep{izenman_multivariate_2008}. 
Motivated by this %
reduced rank assumption, we %
fix the row rank $r$ of our parameter space. %
Each class--specific mean parameter $\MEAN{i}$ satisfies %
\[\MEAN{i} = \lambda_0+\Lambda\alpha_i\xi\;,\]
where $\lambda_0$ is the $b\times \NFEATURES$ matrix of spline coefficients for the population mean functional, the $(\alpha_i)_{i \leq \NCLASSES}$ are the class--specific reduced--rank spline coefficient matrices  of dimensions $r \times r$, and $\Lambda \in \rR^{b \times r}$ and $\xi \in \rR^{r\times \NFEATURES}$  %
allow %
to reconstruct the full--rank spline coefficients for each class from the $\alpha_i$'s. Thus, the final multivariate time--dependent model for $i=1,2,\dots,\NCLASSES$ and $j=1,2,\dots,m_i$ is
\begin{equation}\label{eqn:model}
    Y_{ij}=S_{ij}(\lambda_0+\Lambda\alpha_i\xi+\gamma_{ij})C_{ij}+\epsilon_{ij}\;,
\end{equation}
where %
$\epsilon_{ij}\sim\mathcal{N}_{T_{ij}\times F_{ij}}(0,\sigma^2I,I)$, $\gamma_{ij}\sim\mathcal{N}_{b\times F}(0,\Sigma,\Psi)$, with parameters $\lambda_0$, $\Lambda$, $(\alpha_i)_{i \leq \NCLASSES}$, $\xi$, $\sigma$, $\Sigma$, $\Psi$ and hyperparameters $b$ and $r$ where $r \leq b$.
Equation~\ref{eqn:model} %
handles fragmented curves, misaligned observation times, and accommodates measurement errors without imputation. The resulting decision %
function enables curve classification based on a %
representation of low dimension $r$. Class--specific characteristics are captured by the coefficients $(\alpha_i)_{i \leq \NCLASSES}$. %

However, in this model, $\lambda_0, \Lambda$, $\alpha_i$ and $\xi$'s are interdependent, making parameter estimation harder. To solve this issue, %
we additionally enforce %
the following constraints:
\begin{equation} \label{eqn:constModel}
    \Lambda\text{ and }\xi^\intercal \text{ are orthogonal matrices, \ie }\ \Lambda^\intercal \Lambda=I \quad \text{ and } \quad \xi\xi^\intercal =I \quad \text{ and } \quad \sum_{i=1}^\NCLASSES m_i\alpha_i=0\;.
\end{equation}

\begin{remark}{\textnormal{Extension to FLDA.}}
Equation~\ref{eqn:model} exactly retrieves FLDA when $\NFEATS{i}{j}=|\NFEATURES|=1$ for all $i=1,2,\dots,\NCLASSES$ and $j \in \C{i}$ (without the additional constraints in Equation~\ref{eqn:constModel}). Indeed, in that case, $C=1$, and the class $i$--specific coefficients become $\alpha_i \xi$.
\end{remark}

\begin{remark}{\textnormal{Hyperparameters.}}
The selection of two hyperparameters, $b$ (number of spline functions) and $r$ (parameter rank), has a large impact on the quality of the approximation. We assess the model's performance on different values of $r$ in the experimental study. For the selection of $b$, there is a tradeoff between the richness of the available space of functionals and the computation cost of the inference.
\end{remark}

\subsection{An ECM Algorithm for Parameter Inference}

As previously mentioned, fitting the model described in Equation~\ref{eqn:model} boils down to estimating parameters $\lambda_0,\Lambda, \alpha_i, \xi, \Sigma, \Psi$ and $\sigma$. We describe in Algorithm~\ref{alg:ecm} an expectation/conditional--maximization (ECM) algorithm~\citep{meng1993maximum} for parameter inference. The ECM algorithm is an extension to the well--known expectation/maximization (EM) algorithm for the iterative maximization of the joint log--likelihood on model parameters, which handles missing values in the E--step. This algorithm decouples the inference of most of the parameters across classes $i=1,2,\dots,\NCLASSES$ and samples $j \in \C{i}$. %
If $\gamma_{ij}\sim\mathcal{N}_{b\times c}(0,\Sigma,\Psi)$ then thanks to the properties of matrix--variate normal distributions, $S_{ij}\gamma_{ij}C_{ij}\sim \mathcal{N}_{T_{ij}\times F_{ij}}(0,S_{ij}\Sigma S_{ij}^\intercal ,C_{ij}^\intercal \Psi C_{ij})$. Let us then denote $\Sigma_{ij}' \deff S_{ij}\Sigma S_{ij}^\intercal $ and $\Psi_{ij}' \deff C_{ij}^\intercal \Psi C_{ij}$, which are assumed to be positive definite matrices. Therefore from Equation~\ref{eqn:model}, 
\begin{equation}\label{eq:distr_Y}
vec(Y_{ij})\sim\mathcal{N}_{T_{ij}F_{ij}}(vec(S_{ij}(\lambda_0+\Lambda\alpha_i\xi)C_{ij}), \sigma^2I_{T_{ij}F_{ij}}+\Psi_{ij}'\otimes\Sigma_{ij}')\;.
\end{equation}
In this case, t%
he joint log--likelihood $\ell(\lambda_0, \Lambda, (\alpha_i)_i, \xi, \Sigma, \Psi, \sigma)$ on the parameters can be written as
\begin{eqnarray*}%
        && -\frac{1}{2}\sum_{i=1}^K\sum_{j=1}^{m_i} (vec(Y_{ij}-S_{ij}(\lambda_0+\Lambda\alpha_i\xi)C_{ij})^\intercal (\sigma^2I+\Psi_{ij}'\otimes\Sigma_{ij}')^{-1}vec(Y_{ij}-S_{ij}(\lambda_0+\Lambda\alpha_i\xi)C_{ij}) \\
        && +\log\det(\sigma^2I+\Psi_{ij}'\otimes\Sigma_{ij}')+T_{ij}F_{ij}\log(2\pi))\;. %
\end{eqnarray*}

\begin{algorithm}[htb]
    \caption{MUDRA Algorithm: Model Parameter Inference}%
    \label{alg:ecm}
    \begin{algorithmic}[1]
        \State \textbf{Input:} $\Y{i}{j}$, $S_{ij}$, $C_{ij}$ for $i=1,2,\dots,\NCLASSES$, $j \in \C{i}$
        \State \textbf{Hyperparameters:} $b \geq r$, $r \geq 1$
        \State Initialize at random %
        $\lambda_0$, $\Lambda$, $\alpha_i$, $\xi$, $\Sigma$, $\Psi$, $\sigma$
        \State $Q_0 \gets -\infty \ $, $k \gets 1$
        \State $Q_k \gets \ell(\lambda_0, \Lambda, (\alpha_i)_i, \xi, \Sigma, \Psi, \sigma)$ \textcolor{gray}{\# joint log-likelihood $\ell$}
        \MRepeat%
             \State $k \gets k+1$
            \State Solve the following equations in $\hat{\gamma}_{ij}' \in \rR^{\NTIMES{i}{j} \times \NFEATS{i}{j}}$ for $i \leq \NCLASSES$ and $j \in \C{i}$: 
            \[ \sigma^2(S_{ij}\Sigma S_{ij}^\intercal)^{-1}\hat{\gamma}_{ij}' + \hat{\gamma}_{ij}'C_{ij}^\intercal\Psi C_{ij}=(Y_{ij}-S_{ij}(\lambda_0 + \Lambda\alpha_i\xi)C_{ij})C_{ij}^\intercal\Psi C_{ij} \] %
            \State Solve the following equations in $\hat{\beta}_i \in \rR^{b \times \NFEATURES}$: $\sum_jS_{ij}^\intercal S_{ij}\hat{\beta}_i C_{ij} C_{ij}^\intercal =\sum_jS_{ij}^\intercal (Y_{ij}-\hat{\gamma}'_{ij}) C_{ij}^\intercal$
            \State $\hat{\lambda}_0 \gets \sum_i\frac{m_i\hat{\beta}_i}{\sum_im_i} \ $, $\beta'_i\gets\hat{\beta}_i-\lambda_0 \ $, $\bm{\beta} \gets [\beta'_i, i=1,2,\dots,\NCLASSES] \in \rR^{b\times \NFEATURES \times \NCLASSES}$ %
            \State Find a $r$--way canonical decomposition $(w,\hat{\Lambda},\hat{\xi},\bm{c})$: $\bm{\beta} \approx \sum_{u=1}^r w_u(\hat{\Lambda}_{:,u} \otimes \hat{\xi}_{u,:}^\top \otimes \bm{c}_{u,:})$
            \State $\hat{\alpha}_{i} \gets \sum_{u=1}^r w_u\bm{c}_{i,u}\quad\text{ and }\quad\hat{\gamma}_{ij}\gets(S_{ij}^\intercal S_{ij})^-S_{ij}^\intercal \hat{\gamma}'_{ij}C_{ij}^\intercal (C_{ij}C_{ij}^\intercal )^-$
            \State $\hat{\Sigma}\gets I_b \ $, $\hat{\Psi}\gets I_{\NFEATURES} \ $, $\hat{\sigma} \gets1$
            \While{$\hat{\Sigma}$, $\hat{\Psi}$ and $\hat{\sigma}$ do not converge} %
                \State %
                $\hat{\Sigma}\gets (\sum_{i,j} \NFEATS{i}{j})^{-1} \sum_{i,j}\hat{\gamma}_{ij}\hat{\Psi}^-\hat{\gamma}^\intercal _{ij}\quad\text{ and }\quad \hat{\Psi}\gets (\sum_{i,j}\NTIMES{i}{j})^{-1}\sum_{i,j}\hat{\gamma}^\intercal _{ij}\hat{\Sigma}^-\hat{\gamma}_{ij}$
                \State $\hat{\sigma}^2\gets (\sum_{i,j}T_{ij}\NFEATS{i}{j})^{-1}\sum_{i,j} ||Y_{ij}-S_{ij}(\hat{\lambda}_0+\hat{\Lambda}\hat{\alpha}_i)C_{ij} - \gamma'_{ij}||_F^2+\text{tr}\{ C_{ij}^\intercal \hat{\Psi} C_{ij}\}\text{tr}\{S_{ij}\hat{\Sigma} S_{ij}^\intercal \}$
            \EndWhile
            \State $Q_{k} \gets \ell(\hat{\lambda}_0, \hat{\Lambda}, (\hat{\alpha}_i)_i, \hat{\xi}, \hat{\Sigma}, \hat{\Psi}, \hat{\sigma})$ \textcolor{gray}{\# joint log--likelihood $\ell$ on estimated parameters}
            \If{$Q_{k-1}<Q_{k}$}
            \State $\lambda_0 \gets \hat{\lambda}_0 \ , \Lambda \gets \hat{\Lambda} \ , (\alpha_i)_i \gets (\hat{\alpha}_i)_i \ , \xi \gets \hat{\xi} \ , \Sigma \gets \hat{\Sigma} \ , \Psi \gets \hat{\Psi} \ , \sigma \gets \hat{\sigma}$
            \EndIf
        \EndRepeat
        \State \textbf{until} $Q_{k-1} \geq Q_{k}$\\
        \Return $\lambda_0, \Lambda, (\alpha_i)_i, \xi, \Sigma, \Psi, \sigma$
    \end{algorithmic}
\end{algorithm}

Algorithm~\ref{alg:ecm} aims at maximizing $\ell(\lambda_0, \Lambda, (\alpha_i)_i, \xi, \Sigma, \Psi, \sigma)$ without computing any Kronecker product inverses to remain efficient. It relies on alternatively optimizing each parameter. %
Most prominently, this algorithm features the tensor decomposition algorithm PARAFAC~\citep{bro_parafac_1997} at Line $11$. The Sylvester equation at Line $8$ of Algorithm~\ref{alg:ecm} is solved using the algorithm described in~\cite{bartels_algorithm_1972}. The equation in Line $9$ is solved by executing the generalized minimal residual (GMRES) method~\citep{saad_gmres_1986}. %
Further details are in Appendix \ref{app:ecm}.

\begin{remark}{\textnormal{Theoretical Time Complexity.}}
MUDRA mainly focuses on reliability in the face of missing data for time--series classifications. However, ensuring a tractable algorithm for parameter inference is crucial. In each ECM iteration $k$, there is (1) $\sum_{i \leq \NCLASSES} m_i$ calls to the Bartels--Stewart algorithm (Line $8$), (2) $\NCLASSES$ calls to the GMRES algorithm (Line $9$), (3) one call to the alternating least squares (ALS) minimization procedure for the PARAFAC tensor decomposition at Line $11$, and finally, (4) an iterative optimization step at Lines $14-16$.
\begin{enumerate}
\item One call to the Bartels--Stewart algorithm on sample $j \in \C{i}$ has an overall computational cost in $\mathcal{O}(\NTIMES{i}{j}^3+\NFEATS{i}{j}^3+\NTIMES{i}{j}\NFEATS{i}{j}^2+\NTIMES{i}{j}^2\NFEATS{i}{j})$, by computing the Schur decompositions with the algorithm in~\cite{golub1979hessenberg}.\\
\item GMRES becomes increasingly expensive at each iteration: running $k$ steps of the GMRES method has a time complexity in $\mathcal{O}(b \NFEATURES k^2)$. However, in theory, GMRES converges at step $k \leq b \times \NFEATURES$, so in the worst case, any call to GMRES has a time complexity in  $\mathcal{O}(b^2 \NFEATURES^2)$.\\
\item The PARAFAC decomposition is computed by alternatively minimizing on $w$, $\hat{\Lambda}$, $\hat{\xi}$ and $\bm{c}$. The number of minimization steps across all parameters is limited to $100$ in this function. Each single--parameter minimization is linear in this parameter, and then the time complexity is mainly driven by the matrix inversion and product operations.\\
\item Theoretically, the iterative optimization step in Lines $14-16$ converges, but no convergence rate is provided. In practice, we chose to change the criteria to either stay under a maximum number of iterations ($50$ in our experiments) or until the value of $\hat{\sigma}$ has converged: at step $q$ of the iteration, $|\hat{\sigma}_{q-1}^{-1}(\hat{\sigma}_{q-1}-\hat{\sigma}_q)| < 10^{-5}$. Then, the remaining cost of this step mainly stems from the matrix products.
\end{enumerate}
 We also compare the empirical runtimes of MUDRA and one algorithm from the state--of--the--art, ROCKET~\citep{dempster_rocket_2020} in the experimental study in Section~\ref{sec:exps}.

\end{remark}

\begin{remark}{\textnormal{Implementation.}}
In our implementation of MUDRA, we resort to the Bartels--Stewart algorithm and GMRES present in the Python package \texttt{scipy}~\citep{virtanen2020scipy}. Moreover, we consider the PARAFAC decomposition function in Python package \texttt{Tensorly}~\citep{kossaifi2016tensorly}.
\end{remark}

\subsection{Classification and Dimension Reduction with MUDRA}
\label{sec:class}

Once the parameter values in Equation~\ref{eqn:model} have been infered by running Algorithm~\ref{alg:ecm}, a LDA--like approach to classification (see Equation~\ref{eq:LDA}) can be applied to a new sample $Y \in \rR^{T' \times F'}$. Our model allows to define the density function $f_i$ of class $i=1,2,\dots,\NCLASSES$ and the prior probability $\pi_i$. In our experiments, we assume a uniform prior, that is, $\pi_i = 1/\NCLASSES$ for all $i=1,2,\dots,\NCLASSES$, but other types of prior might also be considered. Then, the class predicted for sample $Y$ is
\begin{eqnarray}\label{eq:MUDRA_classification}
i(Y) \deff \argmax_{i \in \{1,2,\dots,\NCLASSES\}} \log \Prob(C=i | X=Y) \propto \log f_i(Y) + \log\pi_i \;.%
\end{eqnarray}

Let us derive the expression of $f_i$ from Equation~\ref{eq:distr_Y}. Let us denote $S_Y$ the spline matrix in Equation~\ref{eq:Bspline} associated with time points in $Y$, $C_Y$ the corresponding matrix of reordered feature columns, $M_Y \deff \sigma^2I+(C_Y^\intercal \Psi C_Y)\otimes(S_Y\Sigma S_Y^\intercal )$ and 
\[ vec(\hat{\alpha}_Y) \deff \big((\xi C_Y\otimes \Lambda^\intercal S_Y^\intercal )M_Y^-(C_Y^\intercal \xi^\intercal \otimes S_Y\Lambda)\big)^{-\frac{1}{2}}(\xi C_Y\otimes \Lambda^\intercal S_Y^\intercal )M_Y^{-1}vec(Y-S_Y\lambda_0C_Y) \;.\]
Then, ignoring constant factors in $i$ and leveraging the computations made in Appendix~\ref{app:proj}, $\log f_i(Y)$ in Equation~\ref{eq:MUDRA_classification} can be replaced with 
\begin{eqnarray*}
\log f_i(Y) & = & -\frac{1}{2}||vec(Y - S_Y(\lambda_0+\Lambda\alpha_i\xi)C_Y)||^2_{M_Y} + \log\det(M_Y)+T_YF_Y\log(2\pi))\\
& \propto & -\frac{1}{2}||vec(Y - S_Y(\lambda_0+\Lambda\alpha_i\xi)C_Y)||^2_{M_Y}\\
& \propto & -\frac{1}{2}||vec(\hat{\alpha}_Y)-\big((\xi C_Y\otimes \Lambda^\intercal S_Y^\intercal )M_Y^-(C_Y^\intercal \xi^\intercal \otimes S_Y\Lambda)\big)^{\frac{1}{2}}vec(\alpha_i)||\;.
\end{eqnarray*}

Moreover, %
Equation~\ref{eqn:alphaVar} in Appendix~\ref{app:proj} shows that the covariance of $\hat{\alpha}_Y$, $\mathbb{V}[vec(\hat{\alpha}_Y)]=I$. $\hat{\alpha}_Y$ can be reshaped into an $r\times r$ matrix. Then, we obtain an $r^2$--dimensional representation of $Y$. %
\section{Experimental Study}\label{sec:exps}

We describe the empirical results of MUDRA in this section regarding predictive power, interpretability, and time efficiency.

\subsection{Time--Series Classification on a Synthetic Data set}

We built a data set with two features, $12$ time points, two classes, and $300$ samples as described in Appendix~\ref{app:bench}. %
For this experiment, we set $b=7$ %
spline basis functions as a tradeoff between performance and computational cost. %

\begin{figure}[H]
     \centering
     \includesvg[width=0.7\textwidth]{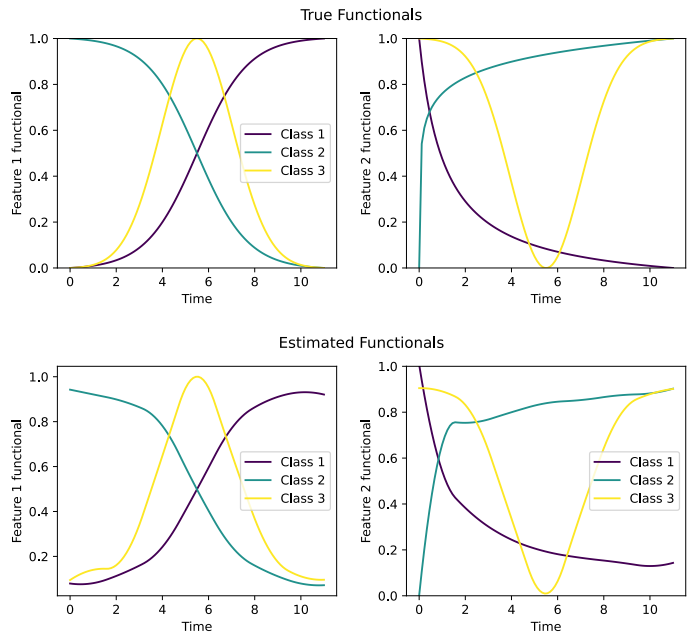}
     \caption{Top plot: curves of feature $1$ (on the left) and feature $2$ (right) across the three classes. Bottom plot: corresponding estimated curves by MUDRA with $r=3$, $b=7$. %
     }
     \label{fig:estim}
 \end{figure}

\begin{figure}[htbp]
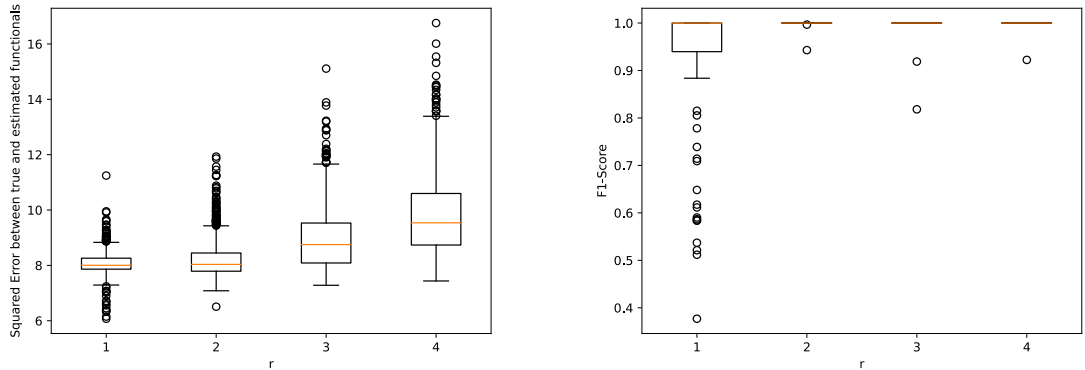

    \centering
    \includesvg[width=0.49\textwidth]{images/SE.svg}
    \hfill
    \includesvg[width=0.49\textwidth]{images/f1_simulated.svg}
    \caption{Synthetic %
    experiments with $b=7$. Left: MSE %
    between estimated and true functional points for %
    $r \in \{1,2,3,4\}$. Right: $F_1$--scores on the classification task on the synthetic data set for $r \in \{1,2,3,4\}$.}%
    \label{fig:mse}
\end{figure}

 To assess the predictive power of our model, we collected the mean squared error (MSE) values between each estimated functional point and its ground truth, and the $F_1$ scores for the corresponding classification task for multiple values of $r$. MSE values and $F_1$ scores are reported in respectively Figure~\ref{fig:mse}a and Figure~\ref{fig:mse}b.
As expected, in Figure~\ref{fig:mse}a, MSE %
decreases when $r$, that is, the representation rank, increases. %
Interestingly, MSE seems to increase again from %
$r=2$, which we interpret as being due to overfitting. Figure~\ref{fig:mse}b shows that for $r>1$, the F1--score for classification is almost 100\%. Thus, the optimal value %
for this data set that balances between performance and speed is $r=2$. %
Figure \ref{fig:estim} compares the true functionals from the synthetic data set and those inferred using MUDRA with hyperparameter $r=3$. %

\subsection{Time--Series Dimension Reduction on a Benchmark Data set}

\begin{figure}
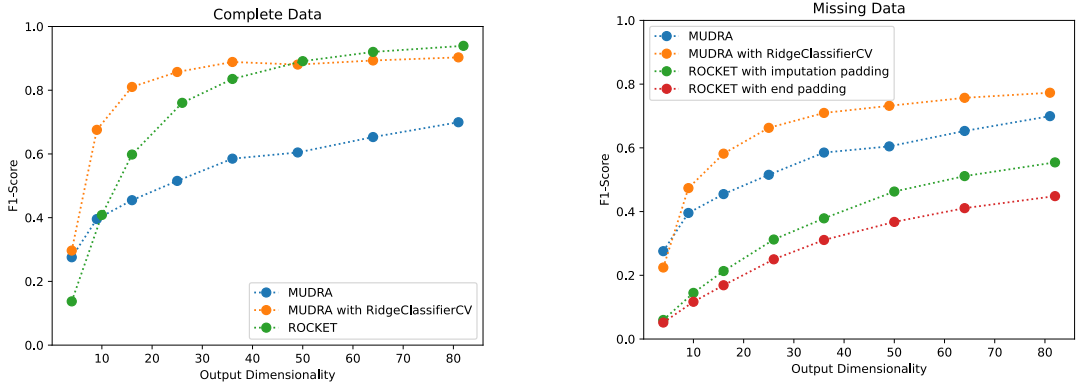

    \centering
    \begin{subfigure}{0.48\textwidth}
        \centering
        \includesvg[width=\textwidth]{images/rocket_vs_mflda_completeData.svg}
        \caption{MUDRA versus ROCKET on complete data. The blue curve correspond to the Bayes--optimal rule in MUDRA, the orange one is the dimension reduction in MUDRA combined with a ridge regression classifier, the green curve correspond to ROCKET--transformed features fed to the classifier.}
        \label{fig:complete}
        \begin{minipage}{1mm}
            \vfill
        \end{minipage}
    \end{subfigure}
    \quad
    \begin{subfigure}{0.48\textwidth}
        \centering
        \includesvg[width=\textwidth]{images/rocket_vs_mflda_missingData.svg}
        \caption{MUDRA versus ROCKET on missing data. The blue curve correspond to the Bayes--optimal rule in MUDRA, the orange one is the dimension reduction in MUDRA combined with a ridge regression classifier, the green (resp., red) curve correspond to ROCKET--transformed features with ``imputation'' (resp., ``end'')--type padding fed to the classifier.}
        \label{fig:missing}
    \end{subfigure}
    \caption{ROCKET versus MUDRA with a ridge regression--based classifier on the ``Articulary Word Recognition'' data set with $b=9$ and for $r \in [1,9]$.}
    \label{fig:word}
\end{figure}

We also applied the MUDRA algorithm %
to the real-world data set ``Articulary Word Recognition'' from \cite{dau_ucr_2019}. Further details about the simulation of missing data are present in %
Appendix~\ref{app:bench}. A recent review by~\cite{ruiz_great_2021} showed that %
ROCKET~\citep{dempster_rocket_2020} is currently the best--performing model, both in terms of efficiency and accuracy, for feature transformation in short--time--series classification. We applied MUDRA or ROCKET to perform a dimension reduction on the data set. Resulting features were fed to %
a ridge regression--based classifier. Such a procedure allows to test how informative the resulting representations are, which is a first step towards interpretability. %
We also classified the data directly through the Bayes--optimal classifier described in %
Section~\ref{sec:class}. 
Since ROCKET always reduces %
to an even number of dimensions, and MUDRA to $r^2$ %
dimensions, to ensure fairness when comparing both algorithms, we considered for ROCKET the smallest even number which was larger than $r^2$. %
The two plots in Figure~\ref{fig:word} report the $F_1$ scores obtained for the classification tasks for multiple values of output dimension $r$ on complete data (\ie without any missing data) and on missing data, where the procedure in Appendix~\ref{app:bench} has been applied to the data set.
Figure \ref{fig:complete} shows that MUDRA competes with ROCKET for very low--dimensional representations, but ROCKET overtakes MUDRA as we increase the output dimension. %
However, when data is missing, ROCKET cannot be applied directly. Inspired by~\cite{bier_variable-length_2022}, we employed two forms of padding to the data set prior to running %
ROCKET:
\begin{enumerate}
    \item \textbf{``Imputation''--type}: The recorded data points are aligned according to their sampling times. %
    Missing time points and missing features are denoted with $0$. %
    \item \textbf{``End''--type}: All missing features are imputed %
    with $0$. All measured %
    time points are concatenated. %
    $0$s are appended at the end to obtain time series of the same length. %
\end{enumerate}

Figure \ref{fig:missing} shows that MUDRA clearly %
outperforms ROCKET in both of these cases.

\subsection{Empirical Runtime of the MUDRA Algorithm}

\begin{figure}[H]
    \centering
    \includesvg[width=0.43\textwidth]{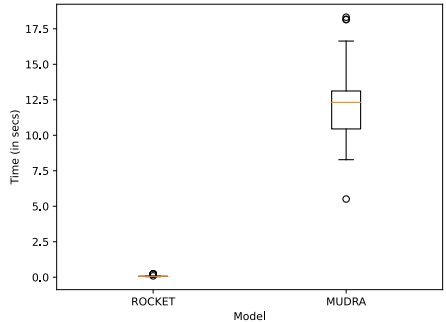}
    \caption{Runtimes for ROCKET and MUDRA on the complete real--life data set for $r=7$ and $b=9$  ($N=1,000$ iterations per algorithm).}
    \label{fig:time}
\end{figure}

To asses the empirical time complexity of MUDRA, we collected the runtimes over $N=1,000$ iterations of both MUDRA and ROCKET combined with a ridge classifier on the real--life data set (without missing data, $r=7$, $b=9$). Resulting runtimes are reported in Figure~\ref{fig:time}.
As evidenced by Figure~\ref{fig:time}, the current implementation of MUDRA is approximately $12$ times slower in average than ROCKET. %
Since ROCKET is a %
simple algorithm that shifts most of the burden of inference to the subsequent classification algorithm, it is extremely fast. However, as previously mentioned, it lacks some of the predictive power of MUDRA.%

\section{Conclusion}

This work presents a novel approach to dealing with missing data in multivariate functional data%
, which has historically proven to be a challenging problem for short--time--series classification. Contemporary algorithms, which achieve optimal performance, typically resort to padding; however, this strategy proves suboptimal for very short time series with a large proportion of missing data. Our proposed algorithm, MUDRA, uniquely enables model training on selective fragments of a high--dimensional surface without necessitating regularization to reconstruct the entire surface for each sample. This eliminates the need for padding and imputation. O%
ur algorithm %
outperforms the current state--of--the--art models in the cases where a very low--dimensional representation is required and/or a large proportion of the data is missing. Such cases frequently arise in real--life data, especially in biomedical applications.

While our approach demonstrates considerable flexibility and efficiency, it has shortcomings. Empirically, we show that the MUDRA algorithm sacrifices some time efficiency in favor of interpretability. Improving on the implementation of Algorithm~\ref{alg:ecm} by resorting to faster routines~\citep{nguyen2016fast,song2022fast}, switching the programming language to C and Fortran, or approximating solution equations~\citep{wang2022inexact} could tackle this issue. 

On the theoretical side, one issue we could not solve was irregularly sampled features for one subject. Consider data being recorded for a sample $\Y{i}{j}$ at time points $\TIME{i}{j}{1}$ and $\TIME{i}{j}{2}$. If the features recorded for $\Y{i}{j}$ at $\TIME{i}{j}{1}$ are not the same as at $\TIME{i}{j}{2}$, our model cannot train on that datapoint without resorting to the imputation or deletion of some of the features. Another significant limitation is the assumption of common covariance matrices $\Sigma$ and $\Psi$ for the auto--regressive noise in each class, which stem from the LDA standpoint. This is sometimes not representative of real data, for example, clinical data, as certain conditions can have a major impact on the variability of levels of certain metabolites. A quadratic discriminant analysis (QDA) approach might have mitigated these issues. However, class--specific covariance matrices raise a number of computational problems in the ECM algorithm and do not lead to the relatively simple expression for the dimension reduction of a new sample. Tackling these shortcomings would improve the application of short--time--series classification algorithms to real--life use cases. %

\begin{acks} %
C.R. has received funding from the European Union's HORIZON 2020 Programme under grant agreement no. 101102016 (RECeSS, HORIZON MSCA Postdoctoral Fellowships--European Fellowships).  O.W. acknowledges support from the German Research Foundation (DFG) FK515800538 (learning convex data spaces).

\subsection*{Conflicts of Interest}
The authors declare that no conflicts of interest were associated with this work.

\subsection*{Code Availability}
The source code for MUDRA (Algorithm~\ref{alg:ecm}), along with documentation, is available at the following GitHub repository: 
\url{https://github.com/rbordoloi/MUDRA} under a GPL--3 license.
\end{acks}

\newpage

\appendix

\section{The ECM Algorithm}

\label{app:ecm}
To guarantee convergence, we assume that $S_{ij}\Sigma S_{ij}^\intercal $ and $C_{ij}^\intercal \Psi C_{ij}$ are invertible for each $i$ and $j$. Our objective is to maximize
 \begin{align*}
    Q =& -\frac{1}{2}\sum_{i,j}\big(\frac{1}{\sigma^2}||Y_{ij}-S_{ij}(\lambda_0 + \Lambda\alpha_i\xi + \gamma_{ij})C_{ij}||^2_F+T_{ij}F_{ij}\log{\sigma^2}+\text{tr}\{\Psi^{-1}\gamma_{ij}^\intercal \Sigma^{-1}\gamma_{ij}'\}\\
    &+ T_{ij}\log\det(\Psi) + F_{ij}\log\det(\Sigma)\big)\;.
\end{align*}
\subsection{The E-Step}\label{sec:estep}
To simplify calculations, we will drop all indices and look at only one data point from one class, as the aggregate can be calculated by doing similar calculations for all data points and computing the arithmetic mean at the end. Thus the simplified model can be written as \[Y = S(\beta+\gamma)C+\epsilon\;,\] where $S$ and $C$ are $T\times b$ and $n\times F$ matrices respectively. 
The idea is to compute the expectation of $S\gamma C=\gamma'$ instead of $\gamma$ itself like in \cite{james_functional_2001} and write $Q$ in terms of $\gamma'$. This will allow us to simplify and use \cite{bartels_algorithm_1972} to find the expectation without computing any tensor products. The log-likelihoods, ignoring proportionality constants, are as follows.
\begin{alignat*}{2}
    l(Y) &= &&-vec(Y-S\beta)^\intercal (\sigma^2I\otimes I + \Psi\otimes\Sigma)^{-1}vec(Y-S\beta C)-\log(|\sigma^2I\otimes I + \Psi\otimes\Sigma)\\
    l(Y,\gamma') &= &&-\frac{vec(Y-S\beta C-\gamma')^\intercal vec(Y-S\beta C-\gamma')}{\sigma^2} - vec(\gamma')^\intercal (( C^\intercal \Psi C)\otimes(S\Sigma S^\intercal ))^{-1}vec(\gamma')\\
    &\,&&- TF\log(\sigma^2) - F\log\det(S\Sigma S^\intercal ) - T\log\det( C^\intercal \Psi C)\\
    &= &&-\frac{1}{\sigma^2}||Y-S\beta C - \gamma'||^2_F-( C^\intercal \Psi C)^{-1}\gamma'^\intercal (S\Sigma S^\intercal )\gamma' - TF\log(\sigma^2)\\
    &\,&&- T\log\det( C^\intercal \Psi C) - F\log\det(S\Sigma S^\intercal )\;.
\end{alignat*}
This is because $\gamma\sim\mathcal{N}_{b\times n}(0,\Sigma,\Psi)\implies\gamma'=S\gamma\sim\mathcal{N}_{m\times n}(0,S\Sigma S^\intercal , C^\intercal \Psi C)$ by Theorem 2.3.10 in \cite{nagar_matrix_1999}. Let us set $\Sigma'=S\Sigma S^\intercal $ and $\Psi'= C^\intercal \Psi C$. We know that $\mathbb{E}[\gamma'|Y=y]=\int\gamma'\frac{f_{Y,\gamma'}(y,\gamma')}{f_Y(y)}\,d\gamma'$, for all $y: f_Y(y)>0$. Let us ignore all terms not depending on $\gamma'$. We will work with the terms in the exponential depending on $\gamma'$ in log space to avoid numerical instability.
\begin{align*}
    &\frac{1}{\sigma^2}(vec(Y-S\beta C)^\intercal vec(\gamma') + vec(\gamma')^\intercal vec(Y-S\beta C) - vec(\gamma')^\intercal vec(\gamma'))\\
    &- vec(\gamma')^\intercal (\Psi'\otimes\Sigma')^{-1}vec(\gamma')\\
    &=\frac{1}{\sigma^2}(vec(Y-S\beta C)^\intercal vec(\gamma') + vec(\gamma')^\intercal vec(Y-S\beta C)\\
    &-vec(\gamma')^\intercal (I+\sigma^2(\Psi'\otimes\Sigma')^{-1})vec(\gamma'))\;.
\end{align*}

Let $M=(I+\sigma^2(\Psi'\otimes\Sigma')^{-1})$. Clearly, $M$ is symmetric. Then by completing the square inside the term multiplied by $-\frac{1}{\sigma^2}$ we have
\begin{align*}
    &-vec(Y-S\beta C)^\intercal M^{-1}vec(Y-S\beta C)+vec(Y-S\beta C)^\intercal M^{-1}vec(Y-S\beta C)\\
    &-vec(Y-S\beta C)^\intercal M^{-\frac{1}{2}}M^{\frac{1}{2}}vec(\gamma')-vec(\gamma')M^{\frac{1}{2}}M^{-\frac{1}{2}}vec(Y-S\beta C)+vec(\gamma')^\intercal Mvec(\gamma')\\
    &=-vec(Y-S\beta)^\intercal M^{-1}vec(Y-S\beta C)\\
    &+(M^{\frac{1}{2}}vec(\gamma')-M^{-\frac{1}{2}}vec(Y-S\beta C))^\intercal (M^{\frac{1}{2}}vec(\gamma')-M^{-\frac{1}{2}}vec(Y-S\beta C))\\
    &=-vec(Y-S\beta C)^\intercal M^{-1}vec(Y-S\beta C)\\
    &+(vec(\gamma')-M^{-1}vec(Y-S\beta C))^\intercal M(vec(\gamma')-M^{-1}vec(Y-S\beta C))\;.
\end{align*}

Clearly, $vec(\gamma')|Y\sim N(M^{-1}vec(Y-S\beta C), \sigma^2M^{-1})$. So
\begin{align*}
    \mathbb{E}[vec(\gamma')|Y]&=M^{-1}vec(Y-S\beta C)\\
    \implies\mathbb{E}[vec(\gamma')|Y]&=(I\otimes I + \sigma^2(\Psi'\otimes\Sigma')^{-1})^{-1}vec(Y-S\beta C)\\
    &=(I\otimes I + (\Psi'^{-1}\otimes I)(I\otimes\sigma^2\Sigma'^{-1}))^{-1}vec(Y-S\beta C)\\
    &=((\Psi'^{-1}\otimes I)(\Psi'\otimes I) + (\Psi'^{-1}\otimes I)(I\otimes\sigma^2\Sigma'^{-1}))^{-1}vec(Y-S\beta C)\\
    &=(\Psi'\otimes I+I\otimes\sigma^2\Sigma'^{-1})^{-1}(\Psi'\otimes I)vec(Y-S\beta C)\\
    &=(\Psi'\otimes I+I\otimes\sigma^2\Sigma'^{-1})^{-1}vec((Y-S\beta C)\Psi'^\intercal )\;.
\end{align*}

Now, we can find $(\Psi'\otimes I+I\otimes\sigma^2\Sigma'^{-1})^{-1}vec((Y-S\beta)\Psi'^\intercal )$ by solving the Sylvester equations $\sigma^2(\Sigma'^{-1})X+X\Psi'^\intercal =(Y-S\beta)\Psi'^\intercal $ using the \cite{bartels_algorithm_1972} algorithm. Let this value be called $\mu_{\gamma'}$. Also, let $\mu_\gamma=\mathbb{E}[\gamma|Y]$ and $Y'=Y-S\beta C$. From \cite{nagar_matrix_1999} we know that is $X\sim\mathcal{N}_{m\times n}(\mu,\Sigma,\Psi)$, $\mathbb{E}[X^\intercal AX]=\text{tr}\{\Sigma A^\intercal \}\Psi + \mu^\intercal A\mu$
\begin{align*}
    \mathbb{E}[||Y'-S\gamma C||_F^2]&=\text{tr}\{\mathbb{E}[(Y'-S\gamma C)^\intercal (Y'-S\gamma C)]\}\\
    &=\text{tr}\{\mathbb{E}[Y'^\intercal Y'+ C^\intercal \gamma^\intercal S^\intercal S\gamma C-Y'^\intercal S\gamma C- C^\intercal \gamma^\intercal S^\intercal Y']\}\\
    &=\text{tr}\{Y'^\intercal Y'\}-2\text{tr}\{Y'^\intercal S\mu_\gamma C\}+\text{tr}\{ C^\intercal \mathbb{E}[\gamma^\intercal S^\intercal S\gamma] C\}\\
    &=\text{tr}\{Y'^\intercal Y'\}-2\text{tr}\{Y'^\intercal S\mu_\gamma C\}+\text{tr}\{ C^\intercal (\text{tr}\{\Sigma S^\intercal S\}\Psi+\mu_\gamma S^\intercal S\mu_\gamma) C\}\\
    &=\text{tr}\{Y'^\intercal Y'\}-2\text{tr}\{Y'^\intercal S\mu_\gamma C\}\\
    &+\text{tr}\{ C^\intercal \text{tr}\{\Sigma S^\intercal S\}\Psi C\}+\text{tr}\{ C^\intercal \mu_\gamma S^\intercal S\mu_\gamma C\}\\
    &=\text{tr}\{Y'^\intercal Y'\}-2\text{tr}\{Y'^\intercal \mu_{\gamma'}\}+\text{tr}\{ C^\intercal \Psi C\}\text{tr}\{S\Sigma S^\intercal \}+\text{tr}\{\mu_{\gamma'}^\intercal \mu_{\gamma'}\}\\
    \implies\mathbb{E}[\sum_{i,j}||Y_{ij}'-S_{ij}\gamma_{ij} C_{ij}||_F^2]&=\sum_{i,j}\Big(||Y_{ij}-S_{ij}(\lambda_0+\Lambda\alpha_i\xi) C_{ij}-\mu_{\gamma_{ij}'}||_F^2+\text{tr}\{ C_{ij}^\intercal \Psi C_{ij}\}\text{tr}\{S_{ij}\Sigma S_{ij}^\intercal \}\Big)\;.
\end{align*}
With some abuse of notation, we will call $\hat{\mu}_{{\gamma'}_{ij}}$ as $\gamma'_{ij}$.
\subsection{The CM-Step}

Since many of the parameters are interdependent, we will use conditional maximization. For a parameter $p$, $\hat{p}$ denotes the estimate at the next step. For convenience we will call $\mu_{\gamma_{ij}'}$ as $\gamma_{ij}'$. Our objective is to maximize
\begin{align*}
    Q =& -\frac{1}{2}\sum_{i,j}\Big(\frac{1}{\sigma^2}\big(||Y_{ij}-S_{ij}(\lambda_0 + \Lambda\alpha_i\xi) C_{ij} - \gamma'_{ij}||^2_F+\text{tr}\{ C_{ij}^\intercal \Psi C_{ij}\}\text{tr}\{S_{ij}\Sigma S_{ij}^\intercal \}\big)+T_{ij}F_{ij}\log{\sigma^2}\\
    &+\text{tr}\{\Psi^{-1}\gamma_{ij}^\intercal \Sigma^{-1}\gamma_{ij}\}+T_{ij}\log\det(\Psi) + F_{ij}\log\det(\Sigma)\}\Big)\;,
\end{align*}
where $||\cdot||_F$ denotes the Frobenius norm.

\subsubsection{Estimating the Linear Parameters}

Equations of the form $\sum_{i}A_iXB_i=C$ for $X$ can be solved by the iterative GMRES algorithm \citep{saad_gmres_1986} without actually computing the inverse of $\sum_{i}(B_i^\intercal \otimes A_i)$. So our task will be to reduce all estimates to this form.

Let $Y_{ij}'=Y_{ij}-\gamma_{ij}'$ and $\beta_i=\lambda_0+\Lambda\alpha_i$. In order to estimate each $\beta_i$, we differentiate $\sum_j||Y_{ij}'-S_{ij}\beta_i C_{ij}||_F^2$ and set to 0.
\begin{align}
    \label{eqn:gmres}
    \begin{split}
        \frac{\partial}{\partial \beta_i}\sum_j\text{tr}\{(Y_{ij}'-S_{ij}\beta_i C_{ij}&)^\intercal (Y_{ij}'-S_{ij}\beta_i C_{ij})\}=0\\
        \implies\sum_j\frac{\partial}{\partial \beta_i}\text{tr}\{ C_{ij}^\intercal \beta_i^\intercal S_{ij}^\intercal S_{ij}\beta_i C_{ij}\}&=2\sum_j\frac{\partial}{\partial \beta_i}\text{tr}\{Y_{ij}'^\intercal S_{ij}\beta_i C_{ij}\}\\
        \implies\sum_j C_{ij} C_{ij}^\intercal \beta_i^\intercal S_{ij}^\intercal S_{ij}&=\sum_j C_{ij}Y_{ij}'^\intercal S_{ij}\\
        \implies\sum_jS_{ij}^\intercal S_{ij}\beta_i C_{ij} C_{ij}^\intercal &=\sum_jS_{ij}^\intercal Y_{ij}' C_{ij}^\intercal\;.
    \end{split}
\end{align}
This is of the form above, so we solve it to estimate each $\beta_i$, say $\hat{\beta}_i$. Then to satisfy Constraint \ref{eqn:constModel} we have,
\begin{equation}
    \label{eqn:demean}
    \hat{\lambda}_0=\sum_i\frac{m_i\hat{\beta}_i}{\sum_im_i}\;.
\end{equation}
Now, let $\beta_i'=\hat{\beta}_i-\hat{\lambda}_0$. We know that $\beta_i'=\hat{\Lambda}\hat{\alpha}_i\hat{\xi}$. Let $\bm{\beta}$ denote the 3-way $b\times F\times K$-tensor formed by stacking $\beta'_i$s. We apply the PARAFAC algorithm \citep{bro_parafac_1997} for $r$ factors on $\bm{\beta}$ to get the $r$-way canonical decomposition, denoted by $\sum_{i=1}^rw_i(\bm{\lambda}_i\otimes\bm{\xi}_i\otimes \bm{c}_i)$, where $\bm{\lambda}_i$, $\bm{\xi}_i$ and $\bm{c}_i$ are normalized. We easily get the estimates of $\hat{\lambda}$ and $\hat{\xi}^\intercal $ subject to Constraint \eqref{eqn:constModel} by stacking the $\bm{\lambda}_i$s and $\bm{\xi}_i$s respectively. Since $\hat{\alpha}_i$s are diagonal, they can be obtained by elementwise multiplication of $w_i$s with $c_i$s, i.e.,
\begin{equation}
    \label{eqn:diagAlphas}
    \hat{\alpha}_{i,jj}=\sum_{j=1}^rw_j\bm{c}_{i,j}\;.
\end{equation}

\subsubsection{Estimating the variance parameters}

$\Sigma$ and $\Psi$ will be estimated by the flip-flop procedure, as suggested by \cite{glanz_expectation-maximization_2013}. Firstly, we need an estimate of $\gamma_{ij}$. We have
\begin{align}
    \label{eqn:gamma}
    \begin{split}
        S_{ij}\gamma_{ij}C_{ij} &= \gamma'_{ij}\\
        \implies(C_{ij}^\intercal \otimes S_{ij})vec(\gamma_{ij}) &= vec(\gamma'_{ij})\\
        \implies vec(\hat{\gamma}_{ij}) &= ((C_{ij}^\intercal \otimes S_{ij})^\intercal (C_{ij}^\intercal \otimes S_{ij}))^-(C_{ij}^\intercal \otimes S_{ij})^\intercal vec(\gamma'_{ij})\\
        \implies vec(\hat{\gamma}_{ij}) &= ((C_{ij}C_{ij}^\intercal )\otimes(S_{ij}^\intercal S_{ij}))^-(C_{ij}\otimes S_{ij}^\intercal )vec(\gamma_{ij}')\\
        \implies vec(\hat{\gamma}_{ij}) &= ((C_{ij}C_{ij}^\intercal )^-\otimes(S_{ij}^\intercal S_{ij})^-)vec(S_{ij}^\intercal \gamma_{ij}'C_{ij}^\intercal )\\
        \implies vec(\hat{\gamma}_{ij}) &= vec((S_{ij}^\intercal S_{ij})^-S_{ij}^\intercal \gamma'_{ij}C_{ij}^\intercal (C_{ij}C_{ij}^\intercal )^-)\\
        \implies\hat{\gamma}_{ij} &= (S_{ij}^\intercal S_{ij})^-S_{ij}^\intercal \gamma'_{ij}C_{ij}^\intercal (C_{ij}C_{ij}^\intercal )^-\;.
    \end{split}
\end{align}

Similar to \cite{glanz_expectation-maximization_2013}, we compute the derivatives

\begin{align*}
    \frac{\partial Q}{\partial\Sigma} &= 0\\
    \implies-\frac{1}{2}\sum_{i,j}\Big(\frac{\text{tr}\{C_{ij}^\intercal \Psi C_{ij}\}}{\sigma^2}S_{ij}^\intercal S_{ij}&-\Sigma^{-1}\gamma_{ij}\Psi^{-1}\gamma_{ij}^\intercal \Sigma^{-1}+F_{ij}\Sigma^{-1}\Big)=0\\
    \implies\Sigma\Big(\sum_{i,j}\frac{\text{tr}\{C_{ij}^\intercal \Psi C_{ij}\}}{\sigma^2}S_{ij}^\intercal S_{ij}\Big)\Sigma&+\Big(\sum_{i,j}F_{ij}\Big)\Sigma=\sum_{i,j}\gamma_{ij}\Psi^{-1}\gamma_{ij}^\intercal \;.
\end{align*}
Directly solving this equation will be computationally intensive, so instead, we ignore the quadratic terms and approximate $\Sigma$ and $\Psi$  by
\begin{equation}
    \label{eqn:cov}
    \hat{\Sigma} = \sum_{i,j}\frac{\hat{\gamma}_{ij}\hat{\Psi}^-\hat{\gamma}^\intercal _{ij}}{\sum_{i,j}F_{ij}},\qquad \text{and}\qquad\hat{\Psi} = \sum_{i,j}\frac{\hat{\gamma}^\intercal _{ij}\hat{\Sigma}^-\hat{\gamma}_{ij}}{\sum_{i,j}T_{ij}}\;.
\end{equation}
Finally, $\sigma^2$ can be estimated as
\begin{equation}
    \label{eqn:sigma2}
    \hat{\sigma}^2=\sum_{i,j}\frac{||Y_{ij}-S_{ij}(\hat{\lambda}_0+\hat{\Lambda}\hat{\alpha}_i)C_{ij} - \gamma'_{ij}||_F^2+\text{tr}\{ C_{ij}^\intercal \hat{\Psi} C_{ij}\}\text{tr}\{S_{ij}\hat{\Sigma} S_{ij}^\intercal \}}{\sum_{i,j}T_{ij}F_{ij}}\;.
\end{equation}After initially estimating the $\hat{\gamma}_{ij}$ values, we proceed to iteratively estimate $\hat{\Sigma}$, $\hat{\Psi}$, and $\hat{\sigma}^2$ until they reach convergence.

\section{Calculations for Projection}
\label{app:proj}
Fix a $Y$, and let $\bm{z} = ((\xi C_Y\otimes \Lambda^\intercal S_Y^\intercal )M_Y^-(C_Y^\intercal \xi^\intercal \otimes S_Y\Lambda))^{-\frac{1}{2}}vec(\hat{\alpha}_Y)=((\xi C_Y\otimes \Lambda^\intercal S_Y^\intercal )M_Y^{-1}(C_Y^\intercal \xi^\intercal \otimes S_Y\Lambda))^-(\xi C_Y\otimes \Lambda^\intercal S_Y^\intercal )M_Y^{-1}vec(Y-S_Y\lambda_0\xi_Y)$, $A=(C^\intercal _Y\xi^\intercal \otimes S_Y\Lambda)$ and $\bm{x}=vec(Y-S_Y\lambda_0C_Y)$. So $\bm{z}=(A^\intercal M_Y^{-1}A)^{-1}A^\intercal M_Y^{-1}\bm{x}$. Then,
\begin{align*}
    &||vec(Y - S_Y\lambda_0C_Y - S_Y\Lambda\alpha_i\xi C_Y)||^2_{M_Y}=||\bm{x} - Avec(\alpha_i)||^2_{M_Y}\\
    =&||\bm{x}-A\bm{z}||^2_{M_Y} + ||\bm{z}-vec(\alpha_i)||^2_{A^\intercal M_Y^{-1}A}+2\langle\bm{x}-A\bm{z},A(\bm{z}-vec(\alpha_i)\rangle_{M_Y}\;,
\end{align*} as $M_Y$ is symmetric positive definite.
Consider only the inner product term. We have,
\begin{align*}
    &\langle\bm{x}-A\bm{z},A(\bm{z}-vec(\alpha_i)\rangle_{M_Y}\\
    ={}&\bm{x}^\intercal M_Y^{-1}A\bm{z}+\bm{z}^\intercal A^\intercal M_Y^{-1}Avec(\alpha_i)-\bm{x}^\intercal M_Y^{-1}Avec(\alpha_i)-\bm{z}^\intercal A^\intercal M_Y^{-1}A\bm{z}\\
    ={}&\bm{x}^\intercal M_Y^{-1}A(A^\intercal M_Y^{-1}A)^{-1}A^\intercal M_Y^{-1}\bm{x}+\bm{x}^\intercal M_Y^{-1}A(A^\intercal M_Y^{-1}A)^{-1}A^\intercal M_Y^{-1}A^\intercal M_Y^{-1}Avec(\alpha_i)\\
    &-\bm{x}^\intercal M_Y^{-1}Avec(\alpha_i)-\bm{x}^\intercal M_Y^{-1}A(A^\intercal M_Y^{-1}A)^{-1}A^\intercal M_Y^{-1}A(A^\intercal M_Y^{-1}A)^{-1}A^\intercal M_Y^{-1}\bm{x}\\
    ={}&\bm{x}^\intercal M_Y^{-1}A(A^\intercal M_Y^{-1}A)^{-1}A^\intercal M_Y^{-1}\bm{x}+\bm{x}^\intercal M_Y^{-1}Avec(\alpha_i)\\
    &-\bm{x}^\intercal M_Y^{-1}Avec(\alpha_i)-\bm{x}^\intercal M_Y^{-1}A(A^\intercal M_Y^{-1}A)^{-1}A^\intercal M_Y^{-1}\bm{x}\\
    \implies&\langle\bm{x}-A\bm{z},A(\bm{z}-vec(\alpha_i)\rangle_{M_Y}=\bm{0}\;.
\end{align*}
Now consider the second norm term. We have,
\begin{align*}
    ||\bm{z}-vec(\alpha_i)||^2_{A^\intercal M_Y^{-1}A}&=||(A^\intercal M_Y^{-1}A)^{-\frac{1}{2}}vec(\hat{\alpha}_Y) - vec(\alpha_i)||^2_{A^\intercal M_Y^{-1}A}\\
    &=||(A^\intercal M_Y^{-1}A)^{-\frac{1}{2}}\big(vec(\hat{\alpha}_Y) - (A^\intercal M_Y^{-1}A)^{\frac{1}{2}}vec(\alpha_i)\big)||^2_{A^\intercal M_Y^{-1}A}\\
    &=||vec(\hat{\alpha}_Y) - (A^\intercal M_Y^{-1}A)^{\frac{1}{2}}vec(\alpha_i)||^2\;.
\end{align*}
Therefore,
\begin{equation}
    \label{eqn:proj}
    ||vec(Y - S_Y\lambda_0C_Y - S_Y\Lambda\alpha_i\xi C_Y)||^2_{M_Y} = ||\bm{x}-A\bm{z}||^2_{M_Y} + ||vec(\hat{\alpha}_Y) - (A^\intercal M_Y^{-1}A)^{\frac{1}{2}}vec(\alpha_i)||^2\;.
\end{equation}

We will also compute the covariance matrix of $vec(\hat{\alpha}_Y)$
\begin{align*}
    \mathbb{V}[\bm{x}] &= cov(vec(Y-S_Y\lambda_0C_Y))\\
    &=\mathbb{V}[vec(S_Y\Lambda\alpha_i\xi C_Y+S_Y\gamma C_Y+\epsilon)]\\
    &=\mathbb{V}[S_Y\gamma C_Y]+\mathbb{V}[\epsilon]\\
    &=(C_Y^\intercal \Psi C)\otimes(S_Y\Sigma S_Y^\intercal )+\sigma^2I=M_Y\\
    vec(\hat{\alpha}_Y) &= (A^\intercal M_Y^{-1}A)^{-\frac{1}{2}}A^\intercal M_Y^{-1}\bm{x}\\
    \implies \mathbb{V}[vec(\hat{\alpha}_Y)]&=(A^\intercal M_Y^{-1}A)^{-\frac{1}{2}}A^\intercal M_Y^{-1}M_YM_Y^{-1}A(A^\intercal M_Y^{-1}A)^{-\frac{1}{2}}\\
    &=(A^\intercal M_Y^{-1}A)^{-\frac{1}{2}}A^\intercal M_Y^{-1}A(A^\intercal M_Y^{-1}A)^{-\frac{1}{2}}\;.
\end{align*}
Thus,
\begin{equation}
    \label{eqn:alphaVar}
    \mathbb{V}[vec(\hat{\alpha}_Y)] = I\;.
\end{equation}

\section[Benchmarking]{Experimental Details}
\label{app:bench}
\subsection[Synthetic Data]{Synthetic Data set}%

To generate simulated data for our study, we selected $F=2$ features, $T=12$ time points, $K=3$ classes, and $m_i=100$ samples per class. A library of six functions with range $[0, 1]$ was created, and each feature-class pair %
was randomly assigned one function. %
The $\Sigma$ (of dimensions ${T\times T}$) and $\Psi$ (of dimensions $F\times F$) matrices were randomly generated, and they were employed to produce autoregressive noise following the matrix-variate normal distribution with a mean matrix of $0$, row covariance $\Sigma$, and column covariance $\Psi$. Additionally, measurement errors were generated by sampling from a matrix-variate normal distribution with a mean matrix of $0$ and identity covariance matrix. Samples were drawn from these matrix-variate normal distributions and added to each selected function $K$ times to generate each sample.

To simulate missing data, for each sample, an integer $T_{ij}$ was randomly sampled from the range of $1$ to $11$, representing the selected number of time points to retain. Similarly, an integer $F_{ij}$ was randomly chosen from the range of $1$ to $2$, indicating the chosen number of features to retain. The $T_{ij}$ timepoints and $F_{ij}$ features were randomly chosen for each sample, and the rest of the data points were deleted.

\subsection[Real Data]{The ``Articulary Word Recognition'' Data set~\citep{ruiz_great_2021}}%

This data set comprises %
quite %
long time series, with $144$ time points. We only retained every $12^{th}$ time point to shorten the length while preserving the overall trend. Then, we randomly deleted some time points and some features in order to simulate the missing data. The proportion of missing time points is capped at $50\%$ whereas it %
is capped at $55\%$ for missing features.

\bibliography{main}

\begin{thebibliography}{36}
\providecommand{\natexlab}[1]{#1}
\providecommand{\url}[1]{\texttt{#1}}
\expandafter\ifx\csname urlstyle\endcsname\relax
  \providecommand{\doi}[1]{doi: #1}\else
  \providecommand{\doi}{doi: \begingroup \urlstyle{rm}\Url}\fi

\bibitem[Bartels and Stewart(1972)]{bartels_algorithm_1972}
R.~H. Bartels and G.~W. Stewart.
\newblock Algorithm 432 [{C2}]: {Solution} of the {Matrix} {Equation} {AX} +
  {XB} = {C} [{F4}].
\newblock \emph{Commun. ACM}, 15\penalty0 (9):\penalty0 820--826, September
  1972.
\newblock ISSN 0001-0782.
\newblock \doi{10.1145/361573.361582}.
\newblock URL \url{https://doi.org/10.1145/361573.361582}.
\newblock Place: New York, NY, USA Publisher: Association for Computing
  Machinery.

\bibitem[Bier et~al.(2022)Bier, Jastrzebska, and
  Olszewski]{bier_variable-length_2022}
Agnieszka Bier, Agnieszka Jastrzebska, and Pawel Olszewski.
\newblock Variable-{Length} {Multivariate} {Time} {Series} {Classification}
  {Using} {ROCKET}: {A} {Case} {Study} of {Incident} {Detection}.
\newblock \emph{IEEE Access}, 10:\penalty0 95701--95715, 2022.
\newblock ISSN 2169-3536.
\newblock \doi{10.1109/ACCESS.2022.3203523}.
\newblock URL \url{https://ieeexplore.ieee.org/document/9874797/}.

\bibitem[Bro(1997)]{bro_parafac_1997}
Rasmus Bro.
\newblock {PARAFAC}. {Tutorial} and applications.
\newblock \emph{Chemometrics and Intelligent Laboratory Systems}, 38\penalty0
  (2):\penalty0 149--171, October 1997.
\newblock ISSN 0169-7439.
\newblock \doi{10.1016/S0169-7439(97)00032-4}.
\newblock URL
  \url{https://www.sciencedirect.com/science/article/pii/S0169743997000324}.

\bibitem[Dau et~al.(2019)Dau, Bagnall, Kamgar, Yeh, Zhu, Gharghabi,
  Ratanamahatana, and Keogh]{dau_ucr_2019}
Hoang~Anh Dau, Anthony Bagnall, Kaveh Kamgar, Chin-Chia~Michael Yeh, Yan Zhu,
  Shaghayegh Gharghabi, Chotirat~Ann Ratanamahatana, and Eamonn Keogh.
\newblock The {UCR} time series archive.
\newblock \emph{IEEE/CAA Journal of Automatica Sinica}, 6\penalty0
  (6):\penalty0 1293--1305, November 2019.
\newblock ISSN 2329-9274.
\newblock \doi{10.1109/JAS.2019.1911747}.
\newblock URL \url{https://ieeexplore.ieee.org/document/8894743}.
\newblock Conference Name: IEEE/CAA Journal of Automatica Sinica.

\bibitem[Dempster et~al.(2020)Dempster, Petitjean, and
  Webb]{dempster_rocket_2020}
Angus Dempster, François Petitjean, and Geoffrey~I. Webb.
\newblock {ROCKET}: exceptionally fast and accurate time series classification
  using random convolutional kernels.
\newblock \emph{Data Min Knowl Disc}, 34\penalty0 (5):\penalty0 1454--1495,
  September 2020.
\newblock ISSN 1573-756X.
\newblock \doi{10.1007/s10618-020-00701-z}.
\newblock URL \url{https://doi.org/10.1007/s10618-020-00701-z}.

\bibitem[Fisher(1936)]{fisher1936use}
Ronald~A Fisher.
\newblock The use of multiple measurements in taxonomic problems.
\newblock \emph{Annals of Eugenics}, 7\penalty0 (2):\penalty0 179--188, 1936.

\bibitem[Fukuda et~al.(2023)Fukuda, Matsui, Takada, Misumi, and
  Konishi]{fukuda_multivariate_2023}
Tatsuya Fukuda, Hidetoshi Matsui, Hiroya Takada, Toshihiro Misumi, and Sadanori
  Konishi.
\newblock Multivariate functional subspace classification for high-dimensional
  longitudinal data.
\newblock \emph{Jpn J Stat Data Sci}, November 2023.
\newblock ISSN 2520-8764.
\newblock \doi{10.1007/s42081-023-00226-x}.
\newblock URL \url{https://doi.org/10.1007/s42081-023-00226-x}.

\bibitem[Gardner-Lubbe(2021)]{gardner-lubbe_linear_2021}
Sugnet Gardner-Lubbe.
\newblock Linear discriminant analysis for multiple functional data analysis.
\newblock \emph{J Appl Stat}, 48\penalty0 (11):\penalty0 1917--1933, 2021.
\newblock ISSN 0266-4763 1360-0532.
\newblock \doi{10.1080/02664763.2020.1780569}.
\newblock Place: England.

\bibitem[Glanz and Carvalho(2013)]{glanz_expectation-maximization_2013}
Hunter Glanz and Luis Carvalho.
\newblock An {Expectation}-{Maximization} {Algorithm} for the {Matrix} {Normal}
  {Distribution}, September 2013.
\newblock URL \url{http://arxiv.org/abs/1309.6609}.

\bibitem[Golub et~al.(1979)Golub, Nash, and Van~Loan]{golub1979hessenberg}
Gene Golub, Stephen Nash, and Charles Van~Loan.
\newblock A hessenberg-schur method for the problem ax+ xb= c.
\newblock \emph{IEEE Transactions on Automatic Control}, 24\penalty0
  (6):\penalty0 909--913, 1979.

\bibitem[Graf et~al.(2024)Graf, Zeldovich, and Friedrich]{graf2024comparing}
Ricarda Graf, Marina Zeldovich, and Sarah Friedrich.
\newblock Comparing linear discriminant analysis and supervised learning
  algorithms for binary classification—a method comparison study.
\newblock \emph{Biometrical Journal}, 66\penalty0 (1):\penalty0 2200098, 2024.

\bibitem[Izenman(2008)]{izenman_multivariate_2008}
Alan~Julian Izenman.
\newblock Multivariate {Regression}.
\newblock In Alan~J. Izenman, editor, \emph{Modern {Multivariate} {Statistical}
  {Techniques}: {Regression}, {Classification}, and {Manifold} {Learning}},
  pages 159--194. Springer New York, New York, NY, 2008.
\newblock ISBN 978-0-387-78189-1.
\newblock \doi{10.1007/978-0-387-78189-1_6}.
\newblock URL \url{https://doi.org/10.1007/978-0-387-78189-1_6}.

\bibitem[James and Hastie(2001)]{james_functional_2001}
Gareth~M. James and Trevor~J. Hastie.
\newblock Functional {Linear} {Discriminant} {Analysis} for {Irregularly}
  {Sampled} {Curves}.
\newblock \emph{Journal of the Royal Statistical Society. Series B (Statistical
  Methodology)}, 63\penalty0 (3):\penalty0 533--550, 2001.
\newblock ISSN 13697412, 14679868.
\newblock URL \url{http://www.jstor.org/stable/2680587}.
\newblock Publisher: [Royal Statistical Society, Wiley].

\bibitem[Jebb et~al.(2015)Jebb, Tay, Wang, and Huang]{jebb_time_2015}
Andrew~T. Jebb, Louis Tay, Wei Wang, and Qiming Huang.
\newblock Time series analysis for psychological research: examining and
  forecasting change.
\newblock \emph{Frontiers in Psychology}, 6, 2015.
\newblock ISSN 1664-1078.
\newblock URL
  \url{https://www.frontiersin.org/journals/psychology/articles/10.3389/fpsyg.2015.00727}.

\bibitem[Kossaifi et~al.(2016)Kossaifi, Panagakis, Anandkumar, and
  Pantic]{kossaifi2016tensorly}
Jean Kossaifi, Yannis Panagakis, Anima Anandkumar, and Maja Pantic.
\newblock Tensorly: Tensor learning in python.
\newblock \emph{arXiv preprint arXiv:1610.09555}, 2016.

\bibitem[Lee and Harring(2023)]{lee_handling_2023}
Daniel~Y. Lee and Jeffrey~R. Harring.
\newblock Handling {Missing} {Data} in {Growth} {Mixture} {Models}.
\newblock \emph{Journal of Educational and Behavioral Statistics}, 48\penalty0
  (3):\penalty0 320--348, June 2023.
\newblock ISSN 1076-9986.
\newblock \doi{10.3102/10769986221149140}.
\newblock URL \url{https://doi.org/10.3102/10769986221149140}.
\newblock Publisher: American Educational Research Association.

\bibitem[Lines et~al.(2016)Lines, Taylor, and Bagnall]{lines_hive-cote_2016}
Jason Lines, Sarah Taylor, and Anthony Bagnall.
\newblock {HIVE}-{COTE}: {The} {Hierarchical} {Vote} {Collective} of
  {Transformation}-{Based} {Ensembles} for {Time} {Series} {Classification}.
\newblock In \emph{2016 {IEEE} 16th {International} {Conference} on {Data}
  {Mining} ({ICDM})}, pages 1041--1046, December 2016.
\newblock \doi{10.1109/ICDM.2016.0133}.
\newblock URL \url{https://ieeexplore.ieee.org/document/7837946}.
\newblock ISSN: 2374-8486.

\bibitem[Mclaughlin and Su(2023)]{mclaughlin2023fedlda}
Connor Mclaughlin and Lili Su.
\newblock Fedlda: Personalized federated learning through collaborative linear
  discriminant analysis.
\newblock In \emph{International Workshop on Federated Learning in the Age of
  Foundation Models in Conjunction with NeurIPS 2023}, 2023.

\bibitem[Meng and Rubin(1993)]{meng1993maximum}
Xiao-Li Meng and Donald~B Rubin.
\newblock Maximum likelihood estimation via the ecm algorithm: A general
  framework.
\newblock \emph{Biometrika}, 80\penalty0 (2):\penalty0 267--278, 1993.

\bibitem[Middlehurst et~al.(2021)Middlehurst, Large, Flynn, Lines, Bostrom, and
  Bagnall]{middlehurst2021hive}
Matthew Middlehurst, James Large, Michael Flynn, Jason Lines, Aaron Bostrom,
  and Anthony Bagnall.
\newblock Hive-cote 2.0: a new meta ensemble for time series classification.
\newblock \emph{Machine Learning}, 110\penalty0 (11-12):\penalty0 3211--3243,
  2021.

\bibitem[Nagar(1999)]{nagar_matrix_1999}
A.~K.~Gupta Nagar, D.~K.
\newblock \emph{Matrix {Variate} {Distributions}}.
\newblock Chapman and Hall/CRC, New York, October 1999.
\newblock ISBN 978-0-203-74928-9.
\newblock \doi{10.1201/9780203749289}.

\bibitem[Nguyen et~al.(2016)Nguyen, Abed-Meraim, and
  Linh-Trung]{nguyen2016fast}
Viet-Dung Nguyen, Karim Abed-Meraim, and Nguyen Linh-Trung.
\newblock Fast adaptive parafac decomposition algorithm with linear complexity.
\newblock In \emph{2016 IEEE International Conference on Acoustics, Speech and
  Signal Processing (ICASSP)}, pages 6235--6239. IEEE, 2016.

\bibitem[{Paul H. C. Eilers} and Marx(1996)]{paul_h_c_eilers_flexible_1996}
{Paul H. C. Eilers} and Brian~D. Marx.
\newblock Flexible {Smoothing} with \${B}\$-splines and {Penalties}.
\newblock \emph{Statistical Science}, 11\penalty0 (2):\penalty0 89--102, 1996.
\newblock ISSN 08834237.
\newblock URL \url{http://www.jstor.org/stable/2246049}.
\newblock Publisher: Institute of Mathematical Statistics.

\bibitem[Ram and Grimm(2009)]{ram_growth_2009}
Nilam Ram and Kevin~J. Grimm.
\newblock Growth {Mixture} {Modeling}: {A} {Method} for {Identifying}
  {Differences} in {Longitudinal} {Change} {Among} {Unobserved} {Groups}.
\newblock \emph{Int J Behav Dev}, 33\penalty0 (6):\penalty0 565--576, 2009.
\newblock ISSN 0165-0254.
\newblock \doi{10.1177/0165025409343765}.
\newblock URL \url{https://www.ncbi.nlm.nih.gov/pmc/articles/PMC3718544/}.

\bibitem[Rao(1948)]{rao1948utilization}
C~Radhakrishna Rao.
\newblock The utilization of multiple measurements in problems of biological
  classification.
\newblock \emph{Journal of the Royal Statistical Society. Series B
  (Methodological)}, 10\penalty0 (2):\penalty0 159--203, 1948.

\bibitem[Ruiz et~al.(2021)Ruiz, Flynn, Large, Middlehurst, and
  Bagnall]{ruiz_great_2021}
Alejandro~Pasos Ruiz, Michael Flynn, James Large, Matthew Middlehurst, and
  Anthony Bagnall.
\newblock The great multivariate time series classification bake off: a review
  and experimental evaluation of recent algorithmic advances.
\newblock \emph{Data Min Knowl Disc}, 35\penalty0 (2):\penalty0 401--449, March
  2021.
\newblock ISSN 1573-756X.
\newblock \doi{10.1007/s10618-020-00727-3}.
\newblock URL \url{https://doi.org/10.1007/s10618-020-00727-3}.

\bibitem[Saad and Schultz(1986)]{saad_gmres_1986}
Youcef Saad and Martin~H. Schultz.
\newblock {GMRES}: {A} {Generalized} {Minimal} {Residual} {Algorithm} for
  {Solving} {Nonsymmetric} {Linear} {Systems}.
\newblock \emph{SIAM J. Sci. and Stat. Comput.}, 7\penalty0 (3):\penalty0
  856--869, July 1986.
\newblock ISSN 0196-5204.
\newblock \doi{10.1137/0907058}.
\newblock URL \url{https://epubs.siam.org/doi/10.1137/0907058}.
\newblock Publisher: Society for Industrial and Applied Mathematics.

\bibitem[Song et~al.(2022)Song, Sebe, and Wang]{song2022fast}
Yue Song, Nicu Sebe, and Wei Wang.
\newblock Fast differentiable matrix square root.
\newblock \emph{arXiv preprint arXiv:2201.08663}, 2022.

\bibitem[Spliid(2016)]{spliid_multivariate_2016}
Henrik Spliid.
\newblock Multivariate {Time} {Series} {Estimation} using marima: 38.
  {Symposium} i {Anvendt} {Statistik} 2016.
\newblock \emph{Symposium i anvendt statistik 2016}, pages 108--123, 2016.
\newblock ISSN 978-87-501-221l-I.
\newblock Publisher: Danmarks Statistik.

\bibitem[Tan et~al.(2022)Tan, Dempster, Bergmeir, and Webb]{tan2022multirocket}
Chang~Wei Tan, Angus Dempster, Christoph Bergmeir, and Geoffrey~I Webb.
\newblock Multirocket: multiple pooling operators and transformations for fast
  and effective time series classification.
\newblock \emph{Data Mining and Knowledge Discovery}, 36\penalty0 (5):\penalty0
  1623--1646, 2022.

\bibitem[Virtanen et~al.(2020)Virtanen, Gommers, Oliphant, Haberland, Reddy,
  Cournapeau, Burovski, Peterson, Weckesser, Bright, et~al.]{virtanen2020scipy}
Pauli Virtanen, Ralf Gommers, Travis~E Oliphant, Matt Haberland, Tyler Reddy,
  David Cournapeau, Evgeni Burovski, Pearu Peterson, Warren Weckesser, Jonathan
  Bright, et~al.
\newblock Scipy 1.0: fundamental algorithms for scientific computing in python.
\newblock \emph{Nature methods}, 17\penalty0 (3):\penalty0 261--272, 2020.

\bibitem[Wang et~al.(2022{\natexlab{a}})Wang, Layton, and
  Barba]{wang2022inexact}
Tingyu Wang, Simon~K Layton, and Lorena~A Barba.
\newblock Inexact gmres iterations and relaxation strategies with
  fast-multipole boundary element method.
\newblock \emph{Advances in Computational Mathematics}, 48\penalty0
  (3):\penalty0 32, 2022{\natexlab{a}}.

\bibitem[Wang et~al.(2022{\natexlab{b}})Wang, Chen, Hershkovich, Yang, Shetty,
  Singh, Jiang, Kotla, Shang, Yerrabelli, Roghanizad, Shandhi, and
  Dunn]{wang_systematic_2022}
Will~Ke Wang, Ina Chen, Leeor Hershkovich, Jiamu Yang, Ayush Shetty, Geetika
  Singh, Yihang Jiang, Aditya Kotla, Jason~Zisheng Shang, Rushil Yerrabelli,
  Ali~R. Roghanizad, Md~Mobashir~Hasan Shandhi, and Jessilyn Dunn.
\newblock A {Systematic} {Review} of {Time} {Series} {Classification}
  {Techniques} {Used} in {Biomedical} {Applications}.
\newblock \emph{Sensors}, 22\penalty0 (20):\penalty0 8016, January
  2022{\natexlab{b}}.
\newblock ISSN 1424-8220.
\newblock \doi{10.3390/s22208016}.
\newblock URL \url{https://www.mdpi.com/1424-8220/22/20/8016}.
\newblock Number: 20 Publisher: Multidisciplinary Digital Publishing Institute.

\bibitem[Yao et~al.(2005)Yao, Müller, and Wang]{yao_functional_2005}
Fang Yao, Hans-Georg Müller, and Jane-Ling Wang.
\newblock Functional {Data} {Analysis} for {Sparse} {Longitudinal} {Data}.
\newblock \emph{Journal of the American Statistical Association}, 100\penalty0
  (470):\penalty0 577--590, 2005.
\newblock ISSN 01621459.
\newblock URL \url{http://www.jstor.org/stable/27590579}.
\newblock Publisher: [American Statistical Association, Taylor \& Francis,
  Ltd.].

\bibitem[Yoon et~al.(2019)Yoon, Zame, and van~der Schaar]{yoon_estimating_2019}
Jinsung Yoon, William~R. Zame, and Mihaela van~der Schaar.
\newblock Estimating {Missing} {Data} in {Temporal} {Data} {Streams} {Using}
  {Multi}-{Directional} {Recurrent} {Neural} {Networks}.
\newblock \emph{IEEE Transactions on Biomedical Engineering}, 66\penalty0
  (5):\penalty0 1477--1490, May 2019.
\newblock ISSN 1558-2531.
\newblock \doi{10.1109/TBME.2018.2874712}.
\newblock URL \url{https://ieeexplore.ieee.org/document/8485748}.
\newblock Conference Name: IEEE Transactions on Biomedical Engineering.

\bibitem[Zhu et~al.(2022)Zhu, Gao, Yang, and Ye]{zhu2022neighborhood}
Fa~Zhu, Junbin Gao, Jian Yang, and Ning Ye.
\newblock Neighborhood linear discriminant analysis.
\newblock \emph{Pattern Recognition}, 123:\penalty0 108422, 2022.

\end{thebibliography}

\end{document}